\documentclass[conference]{IEEEtran}
\IEEEoverridecommandlockouts
\usepackage{cite}
\usepackage{amsmath,amssymb,amsfonts}
\usepackage{algorithmic}
\usepackage{graphicx}
\usepackage{tabularx}
\usepackage{textcomp}
\setdefaultlanguage{english}

\usepackage{booktabs} 
\usepackage{xcolor}
\usepackage{hyperref}
\def\BibTeX{{\rm B\kern-.05em{\sc i\kern-.025em b}\kern-.08em
    T\kern-.1667em\lower.7ex\hbox{E}\kern-.125emX}}

\begin{document}

\title{Tokenization Disparities as Infrastructure Bias: How Subword Systems
Create Inequities in LLM Access and Efficiency
}

\author{
\IEEEauthorblockN{1\textsuperscript{st}Hailay Kidu Teklehaymanot}
\IEEEauthorblockA{\textit{L3S Research Center} \\
\textit{Leibniz University Hannover}\\
Hannover, Germany \\
teklehaymanot@L3S.de}

\and 
\IEEEauthorblockN{2\textsuperscript{nd} Wolfgang Nejdl}
\IEEEauthorblockA{\textit{L3S Research Center} \\
\textit{Leibniz University Hannover}\\
Hannover, Germany \\
nejdl@L3S.de}
}
\maketitle
\begin{abstract}
Tokenization disparities pose a significant barrier to achieving equitable access to artificial intelligence across linguistically diverse populations. This study conducts a large-scale cross-linguistic evaluation of tokenization efficiency in over 200 languages to systematically quantify computational inequities in large language models (LLMs). Using a standardized experimental framework, we applied consistent preprocessing and normalization protocols, followed by uniform tokenization through the \texttt{tiktoken} library across all language samples. Comprehensive tokenization statistics were collected using established evaluation metrics, including Tokens Per Sentence (TPS) and Relative Tokenization Cost (RTC), benchmarked against English baselines. 
Our cross-linguistic analysis reveals substantial and systematic disparities: Latin-script languages consistently exhibit higher tokenization efficiency, while non-Latin and morphologically complex languages incur significantly greater token inflation, often 3--5 times higher RTC ratios. These inefficiencies translate into increased computational costs and reduced effective context utilization for underrepresented languages. Overall, the findings highlight structural inequities in current AI systems, where speakers of low-resource and non-Latin languages face disproportionate computational disadvantages. Future research should prioritize the development of linguistically informed tokenization strategies and adaptive vocabulary construction methods that incorporate typological diversity, ensuring more inclusive and computationally equitable multilingual AI systems.
\end{abstract}

\begin{IEEEkeywords}
Multilingual Models, Tokenization, Infrastructure  Bias, Subword System, Large Language Models 
\end{IEEEkeywords}

\section{Introduction}
Recent advances in large language models (LLMs) have transformed natural language processing \cite{armengol2021multilingual, munoz2021dependency}, yet these developments remain disproportionately concentrated on high-resource languages, particularly English, leaving the majority of the world's languages underrepresented in both research and technological deployment. While multilingual LLMs (mLLMs) theoretically enable cross-lingual knowledge transfer from high-resource to low-resource languages through shared linguistic representations \cite{xu2024survey}, substantial performance disparities persist across linguistically diverse populations \cite{10.1007/978-3-031-41682-8_2}.
Recent analyses of code-switching datasets\cite{dogruoz-etal-2023-representativeness} reveal systemic biases toward English and a lack of sociolinguistic representativeness in data collection and annotation, reflecting broader disparities in multilingual modeling. 
The global NLP research landscape exhibits a significant bias toward high-resource languages, resulting in substantial performance gaps for underrepresented languages due to limited annotated data and inadequate computational support \cite{magueresse2020low, zhu2019importance}.
As highlighted by \cite{dogruoz-etal-2023-representativeness}, the lack of representativeness in English-dominant and unbalanced datasets reflects broader structural inequities in multilingual NLP. These disparities parallel the biases embedded in tokenization systems, underscoring how insufficiently representative data undermines fairness and inclusivity across language technologies. Despite advances in transfer learning and transformer architectures that partially address these disparities through cross-lingual knowledge transfer \cite{theodoropoulos2023information}, the extent and limitations of multilingual generalization remain under investigation. 

A critical but underexplored factor contributing to these disparities is tokenization, the fundamental preprocessing step that transforms raw text into subword units for model input. Tokenization algorithms, predominantly optimized for high-resource languages, may inadequately handle the morphological complexity and structural properties of low-resource languages, thereby exacerbating existing inequalities \cite{mielke2021between}.
This study presents a systematic investigation of tokenization disparities across over 200 languages using the FLORES-200 benchmark. By analyzing state-of-the-art tokenizers, we quantify variations in cross-lingual efficiency and identify tokenization as a key driver of performance inequities in multilingual NLP. Our findings lay the groundwork for developing more equitable tokenization strategies that enhance fairness and inclusivity across diverse linguistic contexts.

\section{Related Work }
\subsection{Background}
Tokenization constitutes a fundamental preprocessing step in natural language processing systems \cite{hassler2006text,toraman2023impact}, transforming raw text into discrete units for model consumption. Contemporary multilingual language models predominantly employ subword tokenization strategies to address vocabulary constraints while maintaining representational efficiency across linguistically diverse corpora.

  \subsection*{Subword Tokenization}
  Subword tokenization decomposes words into semantically meaningful subunits, effectively addressing the open vocabulary problem in neural language models \cite{park2020decoding}. This approach enables efficient processing of both frequent and rare lexical items by representing common words directly while decomposing infrequent terms into familiar subword components \cite{limisiewicz2023tokenization}. Operating at an intermediate granularity between character-level and word-level representations, subword methods capture morphological regularities particularly beneficial for morphologically rich languages \cite{mielke2021between}.
Predominant subword algorithms include Byte Pair Encoding (BPE) \cite{sennrich-etal-2016-neural}, which iteratively merges frequent symbol pairs; WordPiece \cite{song2020fast}, employing probabilistic merging strategies; and SentencePiece \cite{kudo2018sentencepiece}, treating input as raw Unicode sequences without explicit word boundaries. While transformer-based multilingual models rely extensively on these approaches, vocabulary construction often reflects high-resource language dominance, potentially introducing systematic biases for underrepresented languages \cite{toraman2023impact}.

  \subsection*{Tokenization Efficiency Disparities}
  Recent investigations have revealed substantial tokenization disparities across languages, with significant implications for computational equity. \cite{asprovska2024tokenization} demonstrated efficiency variations up to 13-fold between English and other languages across 108 languages, highlighting how existing tokenization schemes systematically favor high-resource languages. \cite{ahia-etal-2023-languages} and \cite{petrov2023language} further documented tokenization length variations reaching 15-fold for semantically equivalent texts, particularly affecting non-Latin scripts and morphologically complex languages.
These disparities carry practical consequences beyond performance metrics, directly impacting computational costs and accessibility. Token-based pricing models in commercial language services result in disproportionately higher usage costs for speakers of underrepresented languages, creating economic barriers to AI access. Proposed solutions include multilingually fair subword algorithms \cite{petrov2023language} and adaptive gradient-based approaches such as MAGNET \cite{ahia2024magnet}, designed to minimize over-segmentation while accommodating diverse morphological and orthographic characteristics.
Multilingual Vocabulary Allocation
Multilingual models face vocabulary allocation challenges, distributing limited vocabulary capacity across multiple languages \cite{zheng2021allocating}. Inductive biases encoded in tokenizers reflect training corpus distributions, often favoring high-resource languages and resulting in suboptimal representation quality for underrepresented scripts \cite{vyas2021optimally, remy2024trans}. Poor vocabulary allocation leads to excessive subword fragmentation, particularly detrimental for word-level tasks requiring accurate morphological segmentation \cite{limisiewicz2023tokenization}.

\section{Methodology}
\subsection{Dataset}
This study employs the FLORES-200 dataset \cite{nllb-24}, providing standardized parallel text across 200 languages representing diverse linguistic families, scripts, and morphological structures. We utilize the devtest split (1,012 sentences per language) to ensure cross-linguistic consistency and eliminate potential data contamination effects.

\subsection{Tokenization Model}






We employ the cl100k \footnote{\href{https://github.com/openai/tiktoken}{OpenAI's tiktoken}}
 base tokenizer via the tiktoken library, representing production-grade Byte Pair Encoding implementations deployed in state-of-the-art large language models, including GPT-3.5 and GPT-4. While computationally efficient, this tokenizer's training data exhibits high-resource language bias, potentially introducing systematic inefficiencies for underrepresented languages.

\section{Experimental Procedure}
For this study, tokenization experiments are conducted using the tiktoken library, which provides access to OpenAI’s production-grade Byte Pair Encoding (BPE) tokenizers. The tiktoken library implements highly efficient tokenization routines specifically designed for transformer-based language models, including those deployed in OpenAIs GPT-3.5 and GPT-4 architectures. In particular, we employ the cl100kbase encoding, a subword vocabulary widely adopted across OpenAI models that supports a broad range of languages and scripts, though it is primarily optimized for high-resource languages.
The experimental procedure consists of the following steps:
\paragraph{Preprocessing and Normalization}
 All text samples from the FLORES-200 devtest set are first normalized using Unicode Normalization Form C (NFC). This normalization step ensures consistency in character representation across languages, which is particularly critical for scripts that allow multiple valid Unicode encodings of the same grapheme cluster. Such normalization is essential to eliminate artifacts that could distort tokenization behavior, especially in non-Latin and morphologically complex languages.\\
\paragraph{Tokenization with tiktoken}
 Following normalization, each sentence is tokenized using the cl100kbase tokenizer provided by tiktoken\footnote{\href{https://github.com/openai/tiktoken}{OpenAI's tiktoken}}.
 This tokenizer employs a data-driven subword segmentation strategy optimized for compression and computational efficiency. However, due to its construction based primarily on high-resource language corpora, it may introduce biases when applied to underrepresented languages that diverge significantly in morphological or orthographic structure.
\paragraph{Tokenization Statistics Extraction}
 After tokenization, we compute and record several key statistics for each sentence, including:
The total number of tokens generated.
The total number of characters processed.
Per language aggregated metrics such as:
Tokens Per Sentence (TPS): the average number of tokens per sentence.
Characters Per Token (CPT): the average number of characters represented by each token.
\paragraph {Cross-Linguistic Efficiency Analysis}
 To quantify cross-linguistic disparities, we calculate the Relative Tokenization Cost (RTC) for each language. This metric compares each language's TPS value to that of English, which serves as the reference baseline. The RTC provides a normalized indicator of tokenization efficiency, allowing us to evaluate how much more or less efficiently each language is tokenized relative to English.
This experimental setup enables a systematic, language-agnostic evaluation of tokenization efficiency across all 200 FLORES languages. By employing a consistent preprocessing pipeline and a single tokenizer configuration, we aim to isolate and quantify the extent of tokenization inefficiencies that may arise from applying widely used subword algorithms to linguistically diverse languages in state-of-the-art multilingual models.

\section{Tokenizer Evaluation Metrics}
To systematically assess tokenization efficiency across diverse languages, we adopt several quantitative metrics that capture both absolute and relative characteristics of tokenized outputs. These metrics allow for cross-linguistic comparisons that highlight disparities introduced by subword segmentation strategies in multilingual language models.

\subsection{Tokens Per Sentence (TPS)}

The \textbf{Tokens Per Sentence (TPS)} metric computes the average number of tokens generated per sentence after tokenization for each language. Formally, for a language $L$ with $N$ sentences:

\[
TPS(L) = \frac{1}{N} \sum_{i=1}^{N} T_i
\]

where $T_i$ denotes the number of tokens in sentence $i$. This metric reflects the granularity of subword segmentation and provides a language-specific view of tokenization density.

\subsection{Characters Per Token (CPT)}

The \textbf{Characters Per Token (CPT)} metric evaluates the average number of characters represented by each token:

\[
CPT(L) = \frac{\sum_{i=1}^{N} C_i}{\sum_{i=1}^{N} T_i}
\]

where $C_i$ is the character count of sentence $i$, and $T_i$ is its corresponding token count. CPT indicates how efficiently the tokenizer compresses character sequences into tokens. Lower CPT values typically reflect finer-grained segmentations, while higher values may suggest more compact tokenization.

\subsection{Relative Tokenization Cost (RTC)}

To directly quantify cross-linguistic disparities, we define the \textbf{Relative Tokenization Cost (RTC)} using English as the baseline reference language. For any language $L$, RTC is computed as:

\[
RTC(L) = \frac{TPS(L)}{TPS(English)}
\]

An RTC value greater than 1 indicates that language $L$ requires more tokens to represent an equivalent sentence compared to English, signaling potential inefficiencies in tokenization for that language. Conversely, an RTC value less than 1 suggests that language $L$ is tokenized more compactly than English.

\subsection{Aggregate Efficiency Comparison}

We conduct aggregate efficiency comparisons across the entire FLORES-200 \cite{costa2022no} language set to quantify global disparities and assess the extent to which low-resource languages are disproportionately affected. This comprehensive analysis facilitates the identification of systematic biases in existing subword tokenization schemes and quantifies linguistic inclusivity across languages with varying resource availability, orthographic systems, and morphological complexity.

\section{Results and Evaluation}

Our comprehensive evaluation encompasses 200+ languages representing diverse script families, including Latin, Arabic, Devanagari, Ethiopic, Cyrillic, and numerous others. The multilingual corpus exhibits substantial imbalance, with Latin-script languages contributing the largest proportion of content, followed by Arabic-script, Devanagari, and Ethiopic language families.
\begin{figure*}[t]
    \centering
    \includegraphics[width=\textwidth]{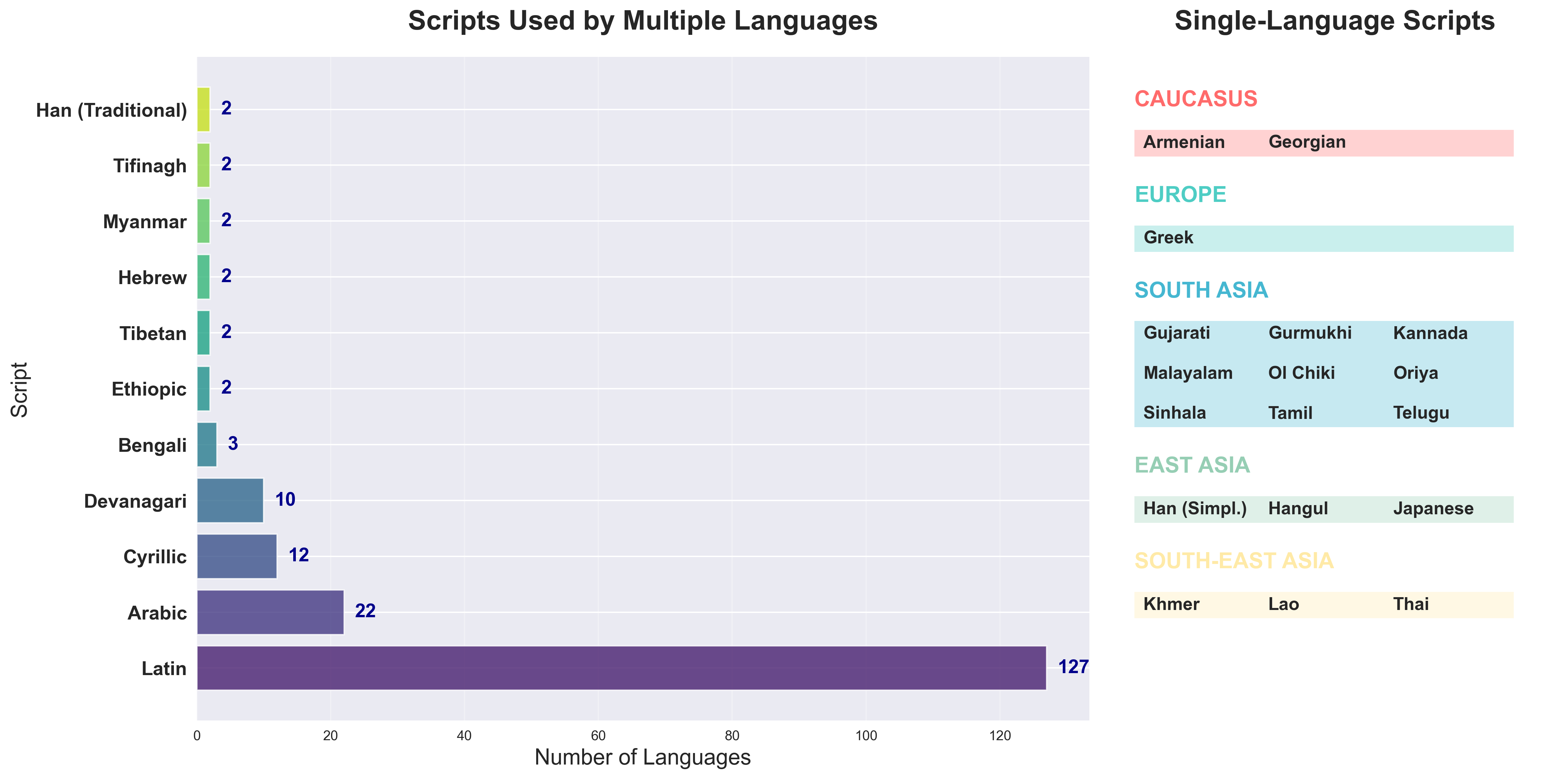}
    \caption{Comparison of scripts used by multiple languages. The bar chart shows the number of languages using each script, with Latin (Latn) being the most widely adopted. An annotation box lists scripts used by only a single language, such as Armenian (Armn), Georgian (Geor), and Japanese (Jpan). This figure highlights the distribution and concentration of script usage across languages.}
    \label{fig:script_distribution}
\end{figure*}

\subsection*{Script Distribution Analysis}
Figure~\ref{fig:script_distribution} illustrates the distribution of languages across script families, revealing significant concentration in Latin-based writing systems. This imbalance reflects historical and contemporary patterns of digital language representation, with implications for tokenization algorithm development and deployment. Scripts employed by single languages including Armenian (Armn), Georgian (Geor), and Japanese (Jpan) represent particularly vulnerable cases where tokenization inefficiencies may disproportionately impact linguistic accessibility.

\subsection*{Cross-Linguistic Tokenization Efficiency}
Our empirical analysis reveals substantial disparities in tokenization efficiency across languages and script families, as demonstrated through multiple complementary analytical perspectives.

\subsection*{Tokens Per Sentence (TPS) Analysis}
\begin{figure}[t]
    \centering
    \includegraphics[width=\columnwidth]{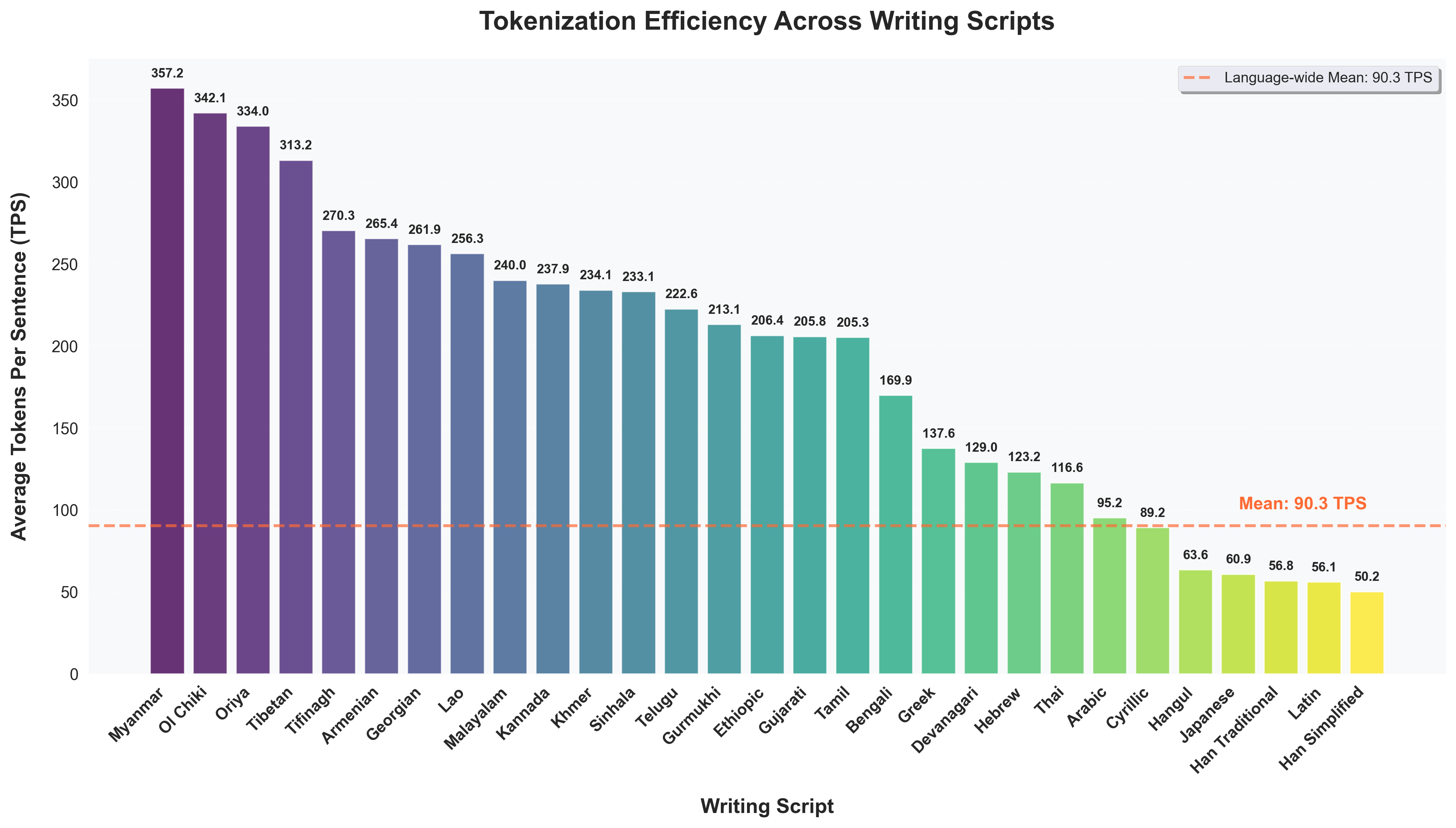}
    \caption{Tokenization Efficiency Across Writing Scripts (Tokens Per Sentence - TPS). Cross-script comparison quantifies extreme tokenization burden variations, with the Myanmar script requiring the highest token density (357.2 TPS) and the Latin script achieving optimal efficiency (50.2 TPS). The language-wide mean of 90.3 TPS serves as a reference, revealing that numerous scripts require 3-4 times more tokens than average, with some scripts demanding nearly 7-fold higher computational costs for equivalent semantic content.}
    \label{fig:script_TPS}
\end{figure}

Figure~\ref{fig:script_TPS} quantifies the extreme range of tokenization burdens across writing systems. Myanmar script requires the highest token density (357.2 TPS), followed by Ol Chiki (342.1 TPS) and Oriya (334.0 TPS), representing nearly 7-fold higher tokenization costs compared to the most efficient scripts. Latin script achieves optimal efficiency (50.2 TPS), with Han Traditional (56.1 TPS) and Han Simplified (56.8 TPS) also demonstrating relatively compact tokenization. The language-wide mean of 90.3 TPS serves as a reference point, revealing that numerous scripts require 3-4 times more tokens than this average, creating substantial computational inequalities.

TPS measurements demonstrate significant variation across languages, with morphologically complex languages exhibiting substantially higher token densities. Languages employing agglutinative morphological processes require extensive subword fragmentation, resulting in elevated TPS values that directly impact computational resource requirements and context window utilization.
\subsection*{Characters Per Token (CPT) Patterns}
\begin{figure}[t]
    \centering
    \includegraphics[width=\columnwidth]{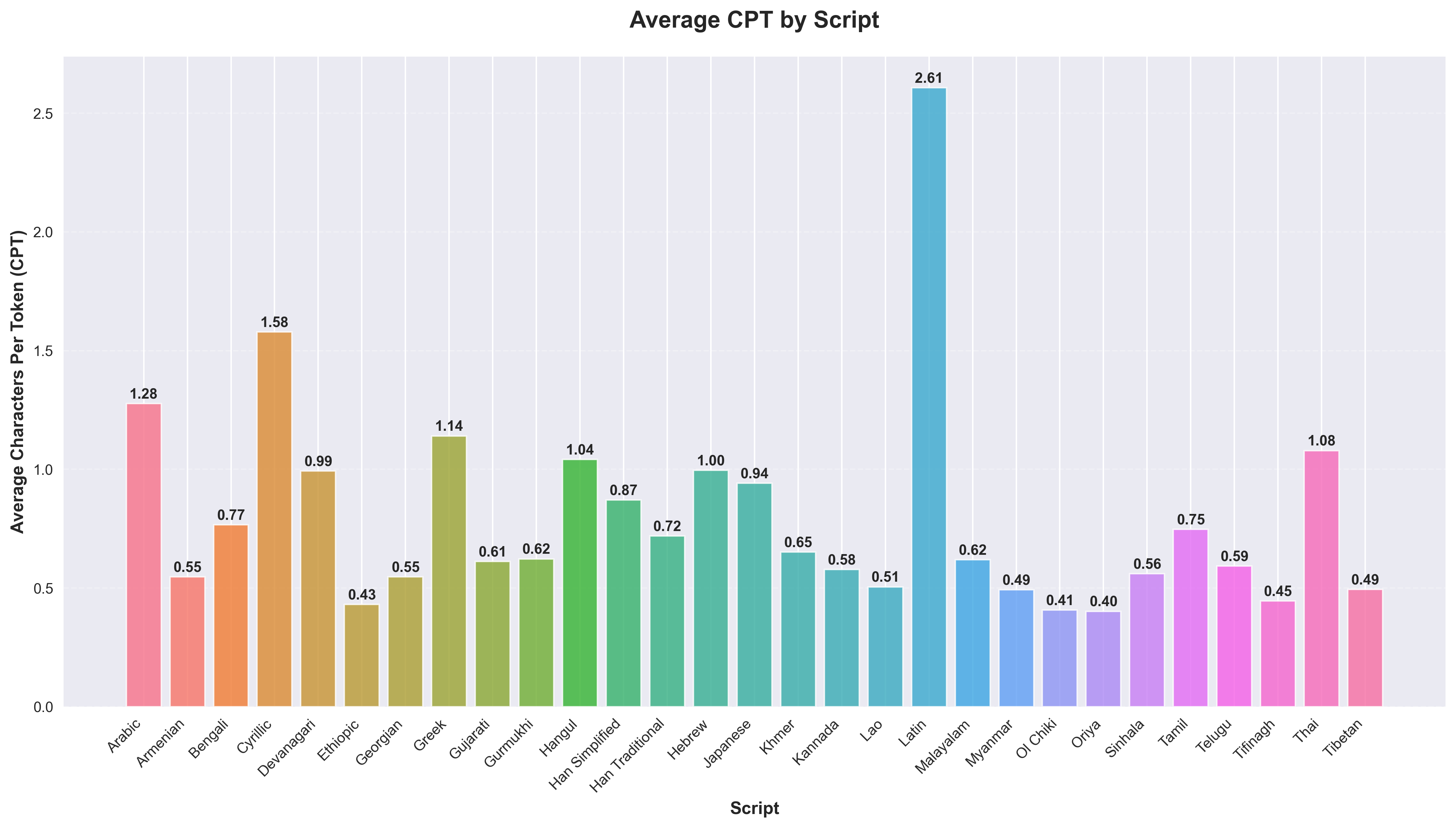}
    \caption{Average Characters Per Token (CPT) by Writing Script. Script-based analysis reveals dramatic efficiency variations, with Latin script achieving optimal compression (2.61 CPT) compared to numerous Asian scripts falling below 1.0 CPT. Non-Latin scripts consistently demonstrate lower efficiency, with Tibetan (0.49 CPT), Oriya (0.40 CPT), and Ol Chiki (0.41 CPT) exhibiting severe tokenization inefficiencies. This pattern reflects the Latin-script optimization inherent in contemporary tokenization algorithms.}
\label{fig:by script}
\end{figure}

	Figure~\ref{fig:by script}
    illustrates dramatic script-dependent variations in compression efficiency. Latin script achieves the highest compression efficiency (2.61 CPT), followed by Cyrillic (1.58 CPT) and Greek (1.14 CPT). Non-Latin scripts consistently demonstrate lower efficiency: Arabic (1.28 CPT), Devanagari (0.99 CPT), and numerous Asian scripts falling below 1.0 CPT. Notably, Tibetan (0.49 CPT), Oriya (0.40 CPT), and Ol Chiki (0.41 CPT) exhibit severe tokenization inefficiencies, indicating excessive subword fragmentation for these writing systems.

\subsection*{Characters Per Token (CPT) Analysis by Language Family}
\begin{figure}[t]
\centering
    \includegraphics[width=\columnwidth]{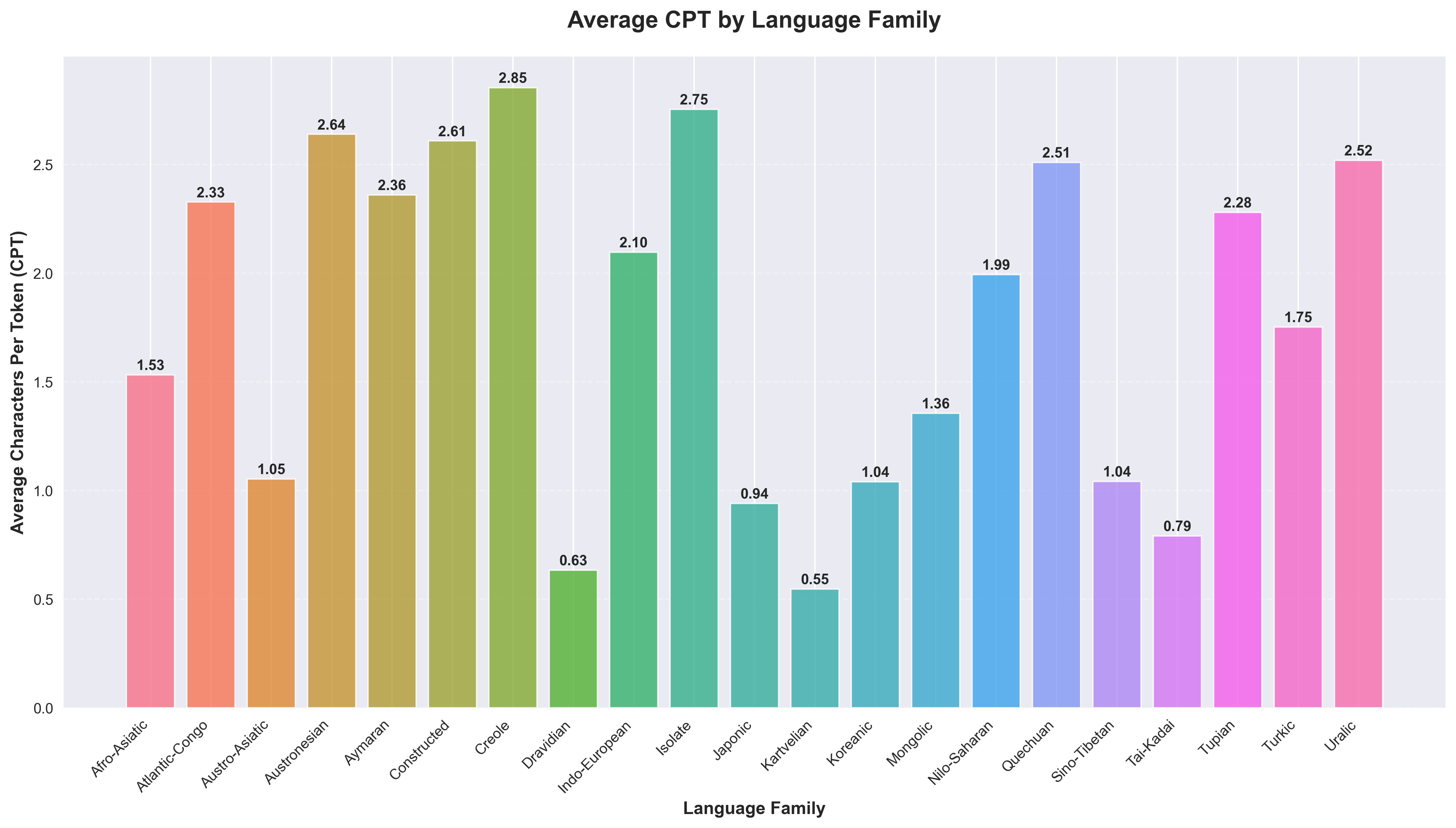}
    \caption{ Average Characters Per Token (CPT) by Language Family. The analysis reveals substantial variation across language families, with Creole languages demonstrating the highest tokenization efficiency (2.85 CPT) and Kannada exhibiting the lowest efficiency (0.55 CPT). Higher CPT values indicate more efficient character-to-token compression, while lower values reflect excessive subword fragmentation. Notable patterns include the superior performance of Creole, Isolate, and Austronesian families compared to morphologically complex families such as Dravidian and Tai-Kadai.}
    \label{fig:script_orgin}
\end{figure}

Figure~\ref {fig:script_orgin} demonstrates substantial variation in CPT values across language families, ranging from 0.55 (Kannada) to 2.85 (Creole languages). Creole languages exhibit the highest compression efficiency (2.85 CPT), followed by Isolate languages (2.75 CPT) and Austronesian languages (2.64 CPT). Conversely, Kannada (0.55 CPT), Dravidian (0.63 CPT), and Tai-Kadai (0.79 CPT) demonstrate substantially lower compression ratios, indicating excessive subword fragmentation that severely impacts computational efficiency for these language families.

\subsection*{Relative Tokenization Cost (RTC) Disparities}
 
RTC measurements quantify the magnitude of cross-linguistic tokenization inequalities relative to English baseline performance. Our results demonstrate RTC values ranging from below 1.0 for closely related Indo-European languages to values exceeding 4.0 for morphologically rich and non-Latin script languages. These disparities translate directly into differential computational costs and accessibility barriers for speakers of underrepresented languages, with some languages requiring up to four times the computational resources for equivalent semantic processing.

\subsection*{Aggregate Efficiency Comparison (AEC)}

The aggregate analysis confirms systematic tokenization biases favoring high-resource, Latin-script languages. Low-resource languages, particularly those employing unique scripts or complex morphological systems, consistently demonstrate reduced efficiency across all evaluation metrics. This pattern reflects the training data composition of contemporary tokenizers and highlights fundamental accessibility challenges in current multilingual AI systems.

\section{Impact on Model Performance and Resource Utilization}
These tokenization disparities carry significant practical implications for multilingual language model deployment:

\paragraph{Computational Resource Inequality} Languages with elevated RTC values require disproportionate computational resources for equivalent semantic processing, creating systematic disadvantages for underrepresented language communities.\\
\paragraph{Context Window Limitations} Higher token densities reduce effective context utilization for complex languages, potentially degrading model performance on tasks requiring extended contextual understanding.
\paragraph{Economic Accessibility Barriers} Token-based pricing models in commercial language services amplify these disparities, resulting in substantially higher usage costs for speakers of inefficiently tokenized languages.
\paragraph{Performance Degradation} Excessive subword fragmentation may impair semantic representation quality, particularly for word-level tasks requiring accurate morphological analysis.

These findings underscore the critical need for developing language-aware tokenization strategies that address systematic inequalities in multilingual AI system accessibility and performance across linguistically diverse populations.

\section{Discussions}
This study frames its findings as evidence of both \textbf{systematic algorithmic} and \textbf{infrastructural biases} rather than mere technical limitations. 
It reveals how existing tokenization algorithms inherently favor high-resource languages, while infrastructural disparities in data availability and tool support further exacerbate the gap. 
The analysis \textbf{connects these technical disparities} to \textbf{broader issues of social and linguistic equity}, emphasizing how such design choices can reinforce \textbf{economic and accessibility barriers}. 
Overall, the study \textbf{underscores the failure of current multilingual systems} to achieve \textbf{inclusive and equitable language representation}.


\section{Conclusions}

Summarizes key findings: 7-fold efficiency disparities
Connects technical inequalities to real-world barriers
Calls for language-aware tokenization strategies
Emphasizes need for equity over efficiency in multilingual AI.

\section*{Limitations}
Acknowledges single tokenizer/dataset constraints
Notes focus on efficiency rather than downstream performance
recognizes potential oversimplification within script families
Provides clear directions for future research.

\section*{Acknowledgment}
This research was supported by the German Academic Exchange Service (DAAD) through the Hilde Domin Programme (funding no. 57615863). The authors gratefully acknowledge this support.

\bibliographystyle{IEEEtran}


\end{document}